# MDD-Thinker: Towards Large Reasoning Models for Major Depressive Disorder Diagnosis

Yuyang Sha[1], Hongxin Pan[1], Gang Luo[1], Caijuan Shi[2], Jing Wang[3*], Kefeng Li[1*]

[1]Center for Artificial Intelligence Driven Drug Discovery, Faculty of Applied Sciences, Macao Polytechnic University, Macau, 999708, Macau SR

[2]College of Artificial Intelligence, North China University of Science and Technology, TangShan 063210, HeBei, China

[3]Department of Critical Care Medicine, Yantai Yuhuangding Hospital, Qingdao University, Yantai 264000, Shandong, China

*Correspondence to J.W ( jinga@qdu.edu.cn) or K.L (kefengl@mpu.edu.mo)

## Summary

**Background**

Major depressive disorder (MDD) is a leading cause of global disability, yet current diagnostic approaches often rely on subjective assessments and lack the ability to integrate multimodal clinical information. Large language models (LLMs) hold promise for enhancing diagnostic accuracy through advanced reasoning but face challenges in interpretability, hallucination, and reliance on synthetic data.

**Methods**

We developed MDD-Thinker, an LLM-based diagnostic framework that integrates supervised fine-tuning (SFT) with reinforcement learning (RL) to strengthen reasoning ability and interpretability. Using the UK Biobank dataset, we generated 40,000 reasoning samples, supplemented with 10,000 samples from publicly available mental health datasets. The model was fine-tuned on these reasoning corpora, and its diagnostic and reasoning performance was evaluated against machine learning, deep learning, and state-of-the-art LLM baselines.

**Findings**

MDD-Thinker achieved an accuracy of 0.8268 and F1-score of 0.8081, significantly outperforming traditional baselines such as SVM and MLP, as well as general-purpose LLMs. Incorporating both SFT and RL yielded the greatest improvements, with relative gains of 29.0% in accuracy, 38.1% in F1-score, and 34.8% in AUC. Moreover, the model demonstrated comparable reasoning performance compared to much larger LLMs, while maintaining computational efficiency.

**Interpretation**

This study presents the first reasoning-enhanced LLM framework for MDD diagnosis trained on large-scale real-world clinical data. By integrating SFT and RL, MDD-Thinker balances accuracy, interpretability, and efficiency, offering a scalable approach for intelligent psychiatric diagnostics. These findings suggest that reasoning-oriented LLMs can provide clinically reliable support for MDD detection and may inform broader applications in mental health care.

**Funding**

This study was funded by Macao Polytechnic University, Science and Technology Development Funds of Macao.

# 1. Introduction

Major depressive disorder (MDD) is one of the most prevalent psychiatric disorders worldwide, characterized by persistent low mood, loss of interest, and cognitive impairments, all of which substantially affect daily life [1]. According to the World Health Organization, depression affects over 280 million people globally and remains a leading cause of disability, posing a major challenge to public health systems worldwide [2]. Early and accurate diagnosis of MDD is critical, as timely intervention can significantly improve treatment outcomes and reduce the risks of chronicity and suicide. Traditionally, MDD diagnosis has relied on standardized clinical criteria, most notably the Diagnostic and Statistical Manual of Mental Disorders (DSM-5) and the International Classification of Diseases (ICD-10/11) [3]. In practice, clinicians frequently employ structured or semi-structured interviews such as the Structured Clinical Interview for DSM Disorders (SCID), as well as validated self-report measures like the Patient Health Questionnaire-9 (PHQ-9) and the Beck Depression Inventory (BDI). While these tools offer a structured framework for symptom assessment, they also face several important limitations [4,5]. First, they rely heavily on subjective self-report, making them vulnerable to recall bias and to variation across cultural and linguistic contexts. Second, their outcomes may vary due to differences in clinician interpretation. Finally, they have limited ability to incorporate multimodal information, such as free-text clinical notes, conversational features, or electronic health records (EHRs). Furthermore, the global disparity in healthcare resources exacerbates these challenges, particularly in low-income and middle-income countries where the scarcity of trained psychiatrists often results in underdiagnosis and delayed treatment.

In recent years, artificial intelligence (AI) has been increasingly explored as a tool for MDD diagnosis [6,7]. Early approaches primarily focused on extracting features from electronic health records (EHRs), psychometric assessments, or speech signals, and subsequently using machine learning or deep learning models to develop MDD diagnostic frameworks. Compared with conventional diagnostic approaches, AI-based methods offer clear advantages by reducing reliance on subjective self-reports, uncovering latent patterns in large-scale datasets, and enabling scalable screening solutions in resource-constrained environments [8]. Large language models (LLMs) have emerged as a promising frontier for advancing MDD diagnostics [9–11]. With their superior capacity for language understanding, LLMs can efficiently process large-scale unstructured medical data, including clinical records, patient self-reports, and online health community content. By leveraging contextual modeling, they are able to capture both common symptoms and subtle semantic nuances that reflect patients' psychological states, while integrating demographic information and medical history to improve generalizability without compromising diagnostic precision [12]. These attributes position LLMs as a strong foundation for consistent, scalable, and sophisticated diagnostic applications in psychiatry. Despite these strengths, several challenges hinder the clinical adoption of LLMs in the diagnosis of MDD. Most existing efforts emphasize classification accuracy while overlooking interpretability, a critical prerequisite for medical trust and decision-making. The opaque "black-box" reasoning of LLMs limits their acceptance among clinicians and patients. Moreover, the widespread reliance on supervised fine-tuning (SFT) primarily enhances instruction-following ability but fails to instill domain-specific medical knowledge, while exacerbating hallucination that pose significant risks in clinical contexts. Furthermore, the scarcity of high-quality annotated data has driven reliance on synthetic datasets, which may introduce bias and lack systematic validation, undermining fairness, reliability, and safety.

To address these challenges, we propose an LLM-based diagnostic framework for MDD, termed MDD-Thinker. The framework incorporates a reinforcement learning paradigm to strengthen the model's reasoning ability, thereby improving interpretability while maintaining diagnostic accuracy. For data construction, we leveraged real-world clinical records from the UK Biobank [13] to generate approximately 50,000 high-quality reasoning samples, further complemented with 10,000 mental health–related information curated from publicly available databases. The training pipeline combines supervised fine-tuning (SFT) with reinforcement learning (RL) to strengthen clinical comprehension, improve diagnostic accuracy for MDD, and enhance reasoning capabilities. To the best of our knowledge, MDD-Thinker represents the first LLM-based framework for MDD diagnosis, integrating reasoning capabilities and trained on large-scale real-world clinical data. We conducted extensive experiments on the UK Biobank and publicly available datasets to evaluate the proposed method. Experimental results indicated that MDD-Thinker consistently surpassed conventional machine learning and deep learning baselines in terms of diagnostic accuracy and consistency. For reasoning capability in MDD diagnosis, our MDD-Thinker 7B demonstrated superior reasoning performance in MDD diagnosis compared with much larger LLMs, including LLaMA 3.1 70B [14] and Qwen 2.5 72B [15]. These findings highlight the efficiency of our approach in balancing performance with computational resources, providing a viable and scalable pathway toward intelligent MDD diagnostics.

## 2. Materials and Methods

### 2.1 Dataset

The UK Biobank [13] is one of the world's most comprehensive biomedical databases, designed to support large-scale epidemiological and genetic research. Between 2006 and 2010, the project recruited over 500,000 participants aged 40–69 across 22 assessment centers in England, Wales, and Scotland. All participants provided electronic informed consent and underwent extensive evaluations, including touchscreen questionnaires, nurse-led interviews, and basic physical examinations. Biological specimens, including blood, urine, and saliva, were collected for biomarker analyses and genotyping, enabling multimodal integration of clinical, phenotypic, and genetic information. In this study, the phenotype for MDD was primarily defined using the ICD-10 code F32, supplemented with self-reported mental health data from UK Biobank fields 20433 and 20434. To ensure data quality, we excluded individuals with more than 30% missing values in the selected features, as well as participants diagnosed with comorbid anxiety or bipolar disorder, in order to minimize potential confounding effects on MDD specificity. After rigorous filtering, a total of 208,406 participants were included, comprising 9,755 individuals diagnosed with MDD and 198,651 controls.

Baseline characteristics of the study cohort are presented in Table 1. To further augment the model's understanding of mental health–related knowledge, an additional set of 10,000 publicly available medical data entries was collected, encompassing multi-turn dialogues, question-answer pairs, and relevant background information. Collectively, these two datasets serve as the basis for generating reasoning datasets, which are subsequently employed during both the SFT and RL stages of model training. The UK Biobank data, together with the collected publicly available datasets, were first curated into reasoning datasets and subsequently used for the model's SFT and RL training phases.

Table 1. The baseline characteristics of participating population.

| Characteristics | HC ($n = 198,651$) | MDD ($n = 9,755$) | $P$ value |
| --- | --- | --- | --- |
| Age, median (IQR) | 61 (53 - 66) | 56 (50 - 63) | < 0.001 |
| BMI, median (IQR) | 26.36 (24.29 - 30.18) | 25.97 (24.13 - 30.30) | 0.047 |
| HDL Cholesterol, median (IQR) | 0.13 (0.61 - 0.33) | 0.13 (0.62 - 0.33) | 0.124 |
| Clinical LDL Cholesterol, median (IQR) | -0.05 (0.64 - 0.70) | -0.05 (0.63 - 0.70) | 0.843 |
| Total Cholesterol, median (IQR) | -0.03 (0.64 - 0.68) | -0.04 (0.62 - 0.67) | 0.664 |
| Triglycerides, median (IQR) | -0.20 (0.42 - 0.62) | -0.20 (0.41 - 0.61) | 0.684 |
| Sleep Duration, median (IQR) | 7 (6 - 8) | 7 (6 - 8) | < 0.001 |
| Sex (%) | | | < 0.001 |
|   Female | 106,497 (53.61) | 5,836 (59.83) | |
|   Male | 92,154 (46.39) | 3,919 (40.17) | |
| Sleeplessness (%) | | | < 0.001 |
|   Usually | 55,543 (27.96) | 3,299 (33.82) | |
|   Sometimes | 95,986 (48.31) | 4,666 (47.83) | |
|   Never | 46,484 (23.40) | 1,787 (18.32) | |
|   Missing Value | 638 (0.21) | 3 (0.03) | |
| Alcohol Frequency (%) | | | < 0.001 |
|   Usually | 44,319 (22.31) | 2,093 (21.46) | |
|   Sometimes | 134,089 (67.50) | 6,703 (68.71) | |
|   Never | 20,143 (10.14) | 934 (9.58) | |
|   Missing Value | 100 (0.05) | 25 (0.25) | |
| Self-harmed Action (%) | | | < 0.001 |
|   Yes | 2,960 (1.49) | 427 (4.38) | |
|   No | 58,264 (29.33) | 5,758 (59.03) | |
|   Not Answer | 238 (0.12) | 2 (0.01) | |
| Missing Value | 137,189 (69.06) | 3,568 (36.58) | |
| Education (%) | | | 0.263 |
|   Low | 38,359 (19.31) | 1,764 (18.08) | |
|   Intermediate | 89,552 (45.08) | 4,396 (45.06) | |
|   High | 65,932 (33.19) | 3,383 (34.68) | |
|   Missing Value | 4,808 (2.42) | 212 (2.18) | |
| Income (%) | | | 0.616 |
|   Low | 34,247 (17.24) | 1,494 (15.32) | |
|   Medium | 126,143 (63.50) | 6,140 (62.95) | |
|   High | 9,714 (4.89) | 530 (5.53) | |
|   Missing Value | 28,547 (14.37) | 1,591 (16.31) | |
| Employ Status (%) | | | 0.617 |
|   Employed | 103,060 (51.88) | 4,815 (49.36) | |
|   Not Employed | 89,313 (44.96) | 4,662 (47.80) | |
|   Other Status | 4,708 (2.37) | 196 (2.01) | |
|   Missing Value | 1,570 (0.79) | 82 (0.83) | |

\* Abbreviations: BMI, body mass index, calculated as weight in kilograms divided by height in meters squared.

Education: low (no relevant qualifications); intermediate (A levels, AS levels, or equivalent; O levels, GCSEs, or equivalent; CSEs or equivalent; NVQ or HND or HNC or equivalent; and other professional qualifications); high (college or university degree).

Income is defined by average total household income before tax: low (less than £18,000); medium (£18,000 to £30,999; £31,000 to £51,999 and £52,000 to £100,000); high (greater than £100,000).

Current employment status: employed (in paid employment or self-employed and full or part-time student); not employed (retired; unable to work because of sickness or disability; unemployed and looking after home and/or family); other status (doing unpaid or voluntary work and none of the above).

For the baseline characteristics including age, BMI, high-density lipoprotein (HDL) cholesterol, clinical low-density lipoprotein (LDL) cholesterol, total cholesterol, triglycerides, and sleep duration, we employed the median and interquartile range (IQR) for statistical representation.

## 2.2 Reasoning Data Synthesis

In this section, we described the synthesis of reasoning data, including feature selection, data filtering, and reasoning path construction. The overall data generation pipeline is illustrated in Figure 1.

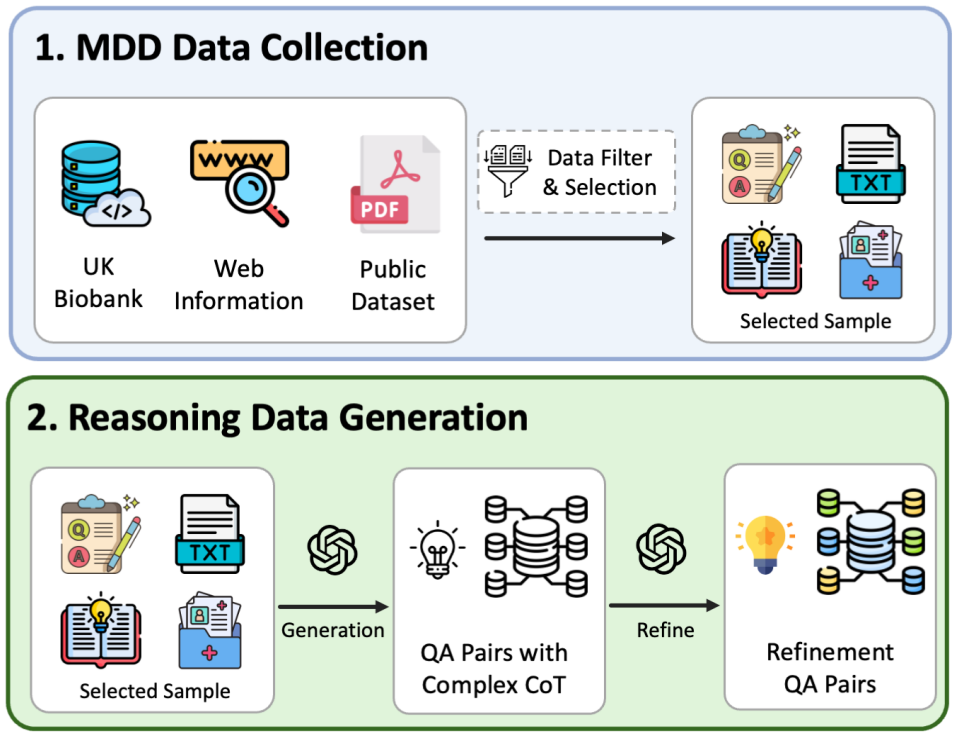

Figure 1. Overall pipeline of reasoning data generation, consisting of two parts: Figure 1.1 illustrates data filtering and selectio`n, while Figure 1.2 presents reasoning data generation and refinement.

**Feature Selection**. Feature selection is an essential step in AI-related works, as the choice of input variables directly affects model performance [16,17]. Although LLM-based methods provide strong prior knowledge and generalization, their effectiveness still depends on the quality and relevance of selected features. In this study, we performed feature selection from two complementary perspectives. First, from the algorithmic perspective, we employed several classical feature ranking methods to identify omics features and biomarkers strongly associated with MDD, which improved the biological interpretability of the model. Second, from the practical perspective, we referred to previous studies [18,19] and consulted psychiatric experts to incorporate additional features related to socioeconomic status, lifestyle, and psychosocial factors. Integrating these two feature selection methods yielded a feature set that is both statistically robust and clinically meaningful, supporting the development of MDD diagnostic and reasoning models. The details of data preprocessing and feature selection process can be found in Appendix S1.

Based on the aforementioned feature selection analysis, we identified 22 variables significantly associated with MDD, encompassing demographic, socioeconomic, lifestyle, psychosocial, clinical, and biochemical factors. Demographic and socioeconomic indicators such as age, sex, education, income, employment status, and length of the working week reflect both biological and environmental influences on mental health. Lifestyle and psychosocial measures, including BMI, sleep patterns, alcohol consumption, history of self-harm or suicidal behavior, perceived happiness, and satisfaction with work, health, family, and finances, capture daily habits and stress-related factors that are closely linked to depressive disorders. Clinical factors, such as long-standing illness, may worsen depressive symptoms, while biochemical markers, including HDL, LDL, total cholesterol, and triglycerides, offer objective measures linked to psychiatric outcomes. These selected features established a robust foundation for fine-tuning the LLM-based MDD diagnostic framework, improving both predictive accuracy and reasoning ability.

**Data Filtering**. Not all UK Biobank samples are suitable for constructing an MDD reasoning dataset, as some may involve misdiagnoses, lack salient features, or present normal clinical indicators. To ensure both data quality and clinical representativeness, we applied a systematic filtering procedure. First, samples with more than 30% missing baseline features were excluded. Second, diagnostic consistency was verified using three advanced LLMs, including GPT-4o [20], Gemini 2.5 [21], and DeepSeek-R1 [22]. Cases that were not correctly classified by all three models were excluded.

**Reasoning Path Construction**. Most existing LLM-based approaches [9,11,23] for MDD diagnosis are typically framed as classification tasks, requiring the model to determine from the input information whether an individual is diagnosed with MDD. However, such direct classification strategy underutilizes the reasoning capacity of LLMs, often leads to hallucinations, and lacks interpretability, limiting clinical trust. Therefore, we developed a novel reasoning path construction method to empower LLM's reasoning capabilities while improving diagnostic accuracy, robustness, and interpretability.

In line with the tabular data processing strategy outlined in the MDD-LLM [9], we reformatted the raw tabular-based data into a coherent textual representation based on predefined rules. For example, the patient's basic information can be expressed in the following form: "*The participant is a 60-year-old female with a body mass index (BMI) of 24.5 kg/m². She experiences occasional sleeplessness and typically sleeps six hours per night. She consumes alcohol about three times per week and has no history of self-harm. She is employed in paid work, earning £45,000 annually, and works 38 hours per week. Her highest education level is O-levels, and she does not have any long-standing illnesses. Clinically, her HDL cholesterol is 2.08 mmol/L, LDL cholesterol is 2.61 mmol/L, total cholesterol is 4.78 mmol/L, and triglycerides are 1.33 mmol/L.*". The details of generation are shown in Appendix S2.

After the process of feature selection and data filtering, we designed a novel data generation method based on Chain-of-Thought (CoT) Prompting and multi-step reasoning to construct reasoning data. It aims to rovide models with explicit, interpretable reasoning paths rather than only final outputs. Specifically, CoT prompting guides the model to first comprehend the input, then decompose complex problems into sequential logical steps. For the multi-step reasoning, the model outputs an intermediate rationale, and the accumulated chain leads to the final conclusion. Specifically, we converted the information of each participant in the UK Biobank into the question–answer pair ($Q$) format. Each question–answer Pair can be defined as $(q_i, a_i) \in Q$, where the $q_i$ and $a_i$ denote the question and corresponding answer, respectively. Then, we incorporated CoT Prompt information ($P$) into question–answer pair, which as be denoted as $(P, q_i, a_i)$. For these above data, we employed GPT-4o to construct reasoning paths ($r_i$) along with the corresponding predicted outcomes ($a_i^*$). And the new question–answer pair can be defined as $(P, q_i, r_i, a_i^*)$. A reasoning path is defined as valid when $a_i^* = a_i$, and invalid when it does not. To enhance the probability of obtaining valid reasoning paths, the model is allowed up to $T$ attempts.

Samples that fail to yield a correct result after these attempts are discarded. Besides, an optimization procedure was employed to further refine the reasoning data. First, each original sample $(P, q_i, r_i)$ was fed into the model to refine the CoT Prompting $P$, yielding an enhanced prompt $P^*$ that provides stronger guidance for multi-step reasoning. Subsequently, we took $(P^*, q_i)$ as input and generated a revised reasoning path $\hat{r}_i^*$ along with its corresponding answer $\hat{a}_i^*$. To ensure both completeness and interpretability, the model explicitly produced intermediate steps, forming a coherent and transparent logical chain. If $\hat{a}_i^*$ matched the ground truth $a_i$, the new generated reasoning path $\hat{r}_i^*$ was retained. Otherwise, the data generated framework repeated the generation process up to $N$ times. Samples failing after $N$ attempts were reverted to the original generated pairs $(P, q_i, r_i, a_i)$ to preserve dataset reliability and quality.

**Feature Data Collection**. For data collection, in addition to the dataset specifically generated for this study, we incorporated a diverse set of publicly available psychological datasets covering a broad spectrum of mental health–related knowledge. These external resources not only expanded the breadth of information available to the model but also enhanced the robustness of its reasoning capabilities across varied clinical scenarios. To ensure consistency and reliability, the selected public datasets were preprocessed and refined following the same systematic procedure. As a result, we obtained approximately 40K high-quality reasoning samples derived from the UK Biobank and an additional 10K samples generated from public datasets, collectively forming a comprehensive and representative corpus for subsequent model training and evaluation.

### 2.3 MDD-Thinker Fine-tuning

We adopted a two-stage training approach for MDD-Thinker: Supervised Fine-Tuning (SFT) and Reinforcement Learning (RL). SFT enabled the model to learn domain-specific knowledge and initial reasoning from curated reasoning corpora. RL guided the model via rewards to produce accurate, logically coherent, and interpretable outputs. An overview of the training pipeline is provided in Figure 2, while detailed descriptions of each stage are given in the following sections.

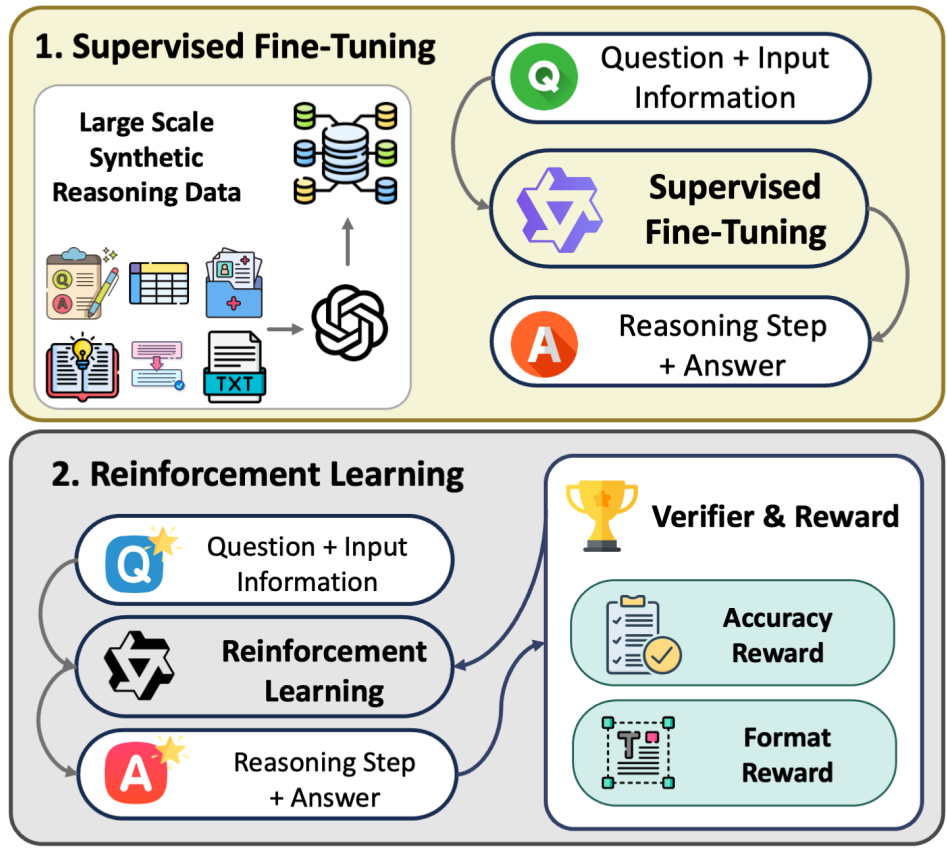

Figure 2. Training pipeline of the proposed MDD-Thinker, comprising two stages: Figure 2.1 illustrates Supervised Fine-Tuning (SFT), and Figure 2.2 illustrates Reinforcement Learning (RL).

**SFT Stage**. SFT constitutes a critical stage in adapting LLMs for domain-specific tasks. In this study, SFT was employed to equip the proposed method with fundamental reasoning capabilities tailored for MDD diagnosis. Specifically, we initialized from the general-purpose LLM and fine-tuned it on the generated SFT dataset, which incorporates both synthetic reasoning samples and selected external corpora related to mental health. This design ensures that the model not only captures domain knowledge but also learns to follow structured reasoning patterns. During training stage, we adopted a fully parameterized SFT scheme, where all model parameters were updated to maximize adaptation efficiency. Consistent with standard practice in autoregressive language modeling, the model was trained to predict the next token given preceding context, thereby learning to generate coherent reasoning trajectories aligned with clinical logic. The optimization objective was the minimization of the negative log-likelihood over the target sequence, formally expressed as:

$$\mathcal{L}_{SFT}(\theta) = -\sum_{t=1}^{T} \log P_\theta(y_t | y_{1:t-1}, x) \quad (1)$$

where $x$ denotes the input information, $y_t$ is the ground-truth token at position $t$, $y_{1:t-1}$ represents the preceding tokens, and $\theta$ are the model parameters.

**RL Stage**. In this study, to further enhance the performance of LLM-based approaches in MDD diagnosis and reasoning tasks, we introduced a RL stage following SFT. Specifically, we employed GRPO [24] to refine the model, aiming to improve not only diagnostic accuracy but also the logical coherence and interpretability of its reasoning process. For each input $q \in P(Q)$, the model generates $G$ candidate reasoning paths $\{o_i\}_{i=1}^{G}$ under the old policy $\pi_\theta^{old}$, The updated policy $\pi_\theta^{new}$ is optimized by maximizing the expected clipped advantage while regularizing the KL divergence with a reference policy $\pi_{ref}$. Each output is evaluated using a reward function combining accuracy $r_{acc}$ and format constraints $r_{fmt}$. The GRPO objective can be defined as follows:

$$\mathcal{L}_{GRPO}(\theta) = \mathbb{E}_{q,\{o_i\}\sim\pi_{old}} \frac{1}{G} \sum_{i=1}^{G} [\min(r_i, A_i, \text{clip}(r_i, 1-\varepsilon, 1+\varepsilon)A_i) - \beta \mathbb{D}_{KL}(\pi_\theta^{new} || \pi_{ref})] \quad (2)$$

$$r_i = \frac{\pi_\theta^{new}(o_i|q)}{\pi_\theta^{old}(o_i|q)} A_i \quad (3)$$

$$r(o) = \mu \cdot r_{acc}(o) + \nu \cdot r_{fmt}(o) \quad (4)$$

where $A_i$ is the estimated advantage, and $\varepsilon$, $\beta$, $\mu$, $\nu$ are hyperparameters controlling clipping and reward weighting.

## 2.4 Implement Details

We adopted Qwen2.5 7B [15] as the base model, implemented in Python with PyTorch, and trained on eight NVIDIA H800 GPUs. During the SFT stage, the model was trained on the designed SFT dataset for 3 epochs, using a learning rate of $1 \times 10^{-5}$ and a batch size of 256. In the subsequent RL stage, training was performed on the generated

RL dataset for 2 epochs with a learning rate of $1 \times 10^{-6}$ and the a batch size of 256. Within the GRPO framework, each query generated 8 candidate outputs, which facilitated more stable reward estimation and policy optimization.

## 3. Experiments and Results

### 3.1 Experimental Setting

In this paper, we primary focused on evaluating the diagnostic performance and reasoning ability of the proposed MDD-Thinker. To this end, we first compared the proposed method with commonly used machine learning and deep learning models to assess diagnostic capability. The baseline models include SVM, RF, XGBoost, MLP, ResNet1D [25], and MDD-LLM [9]. Furthermore, we compared our designed approach with representative LLMs to evaluate both diagnostic and reasoning abilities. The selected LLM baselines compared LLaMA3.1 8B [14], Qwen2.5 7B [15], Qwen2.5 72B [15] and GPT-4o [20].

### 3.2 Evalution Metrics

In order to evaluate diagnostic accuracy for MDD, we adopted a comprehensive set of metrics to rigorously assess model performance. For the MDD diagnostic, we employed Accuracy, Precision, Recall, F1-score, and AUROC to evaluate diagnostic ability. To ensure statistical rigor in model comparison, we further applied the DeLong test to assess the significance of differences in AUC, thereby validating the reliability of observed performance improvements. In addition, we also evaluated the quality of LLM-based methods generated outputs with BLEU, ROUGE, and METEOR. Detailed definitions and descriptions of these evaluation metrics are provided in Appendix S3.

### 3.3 Results on UK Biobank

The experimental results demonstrate that the proposed MDD-Thinker model substantially outperforms traditional machine learning methods, deep learning architectures, and publicly available LLM-based approaches in diagnosing MDD. As shown in Table 2, MDD-Thinker achieved an accuracy of 0.8268 and an F1-score of 0.8081. Compared with the classical machine learning baseline SVM, our model yielded relative improvements of approximately 21.95% in Accuracy and 24.28% in F1-score. When benchmarked against the MLP, MDD-Thinker achieved an relative gain of 20.67% in Accuracy and 28.53% in F1-score, underscoring the advantage of reasoning-enhanced fine-tuning over conventional deep learning architectures. The inferior performance of publicly released LLMs can be attributed to their lack of domain-specific adaptation to MDD datasets. However, our porposed method built upon the Qwen2.5-7B and further fine-tuned with domain-relevant data, consistently delivered superior diagnostic performance across all evaluation metrics. To further illustrate model effectiveness, Figure 3 presents a comparison of ROC curves. The DeLong test produced a p-value of 0.0031, far below 0.05. This finding shows that MDD-Thinker provides significant improvements in diagnostic accuracy compared with baseline methods.

Table 2. Comparison results of different methods on the UK Biobank dataset.

| Method | ACC | F1 | AUC | SPE | SENS | PPV | NPV |
|---|---|---|---|---|---|---|---|
| SVM | 0.6794 | 0.6517 | 0.7463 | 0.6958 | 0.6596 | 0.6438 | 0.7106 |
| RF | 0.6883 | 0.6341 | 0.7369 | 0.7662 | 0.5947 | 0.6791 | 0.6943 |
| LightGBM | 0.7091 | 0.6707 | 0.7693 | 0.7572 | 0.6516 | 0.6908 | 0.7231 |
| XGBoost | 0.7068 | 0.6704 | 0.7497 | 0.7501 | 0.6554 | 0.6858 | 0.7233 |
| CatBoost | 0.7117 | 0.6717 | 0.7751 | 0.7642 | 0.6486 | 0.6961 | 0.7232 |
| MLP | 0.6869 | 0.6301 | 0.7522 | 0.7692 | 0.5877 | 0.6793 | 0.6916 |
| ResNet1D | 0.7077 | 0.6644 | 0.7654 | 0.7669 | 0.6369 | 0.6945 | 0.7173 |
| LLaMA3.1 8B | 0.6167 | 0.5229 | 0.6387 | 0.7452 | 0.4625 | 0.6013 | 0.6249 |
| Qwen2.5 7B | 0.6409 | 0.5852 | 0.6532 | 0.7593 | 0.5411 | 0.6182 | 0.6483 |
| MDD-LLM 8B | 0.7919 | 0.7642 | 0.8579 | 0.8039 | 0.7763 | 0.7524 | 0.8241 |
| MDD-Thinker 7B | 0.8268 | 0.8081 | 0.8803 | 0.8229 | 0.8291 | 0.7838 | 0.8614 |

* ACC = accuracy, F1 = f1 score, AUC = area under the receiver-operating characteristic curve,

SPE = specificity, SEN = sensitivity, PPV = positive predictive value, NPV = negative predictive value.

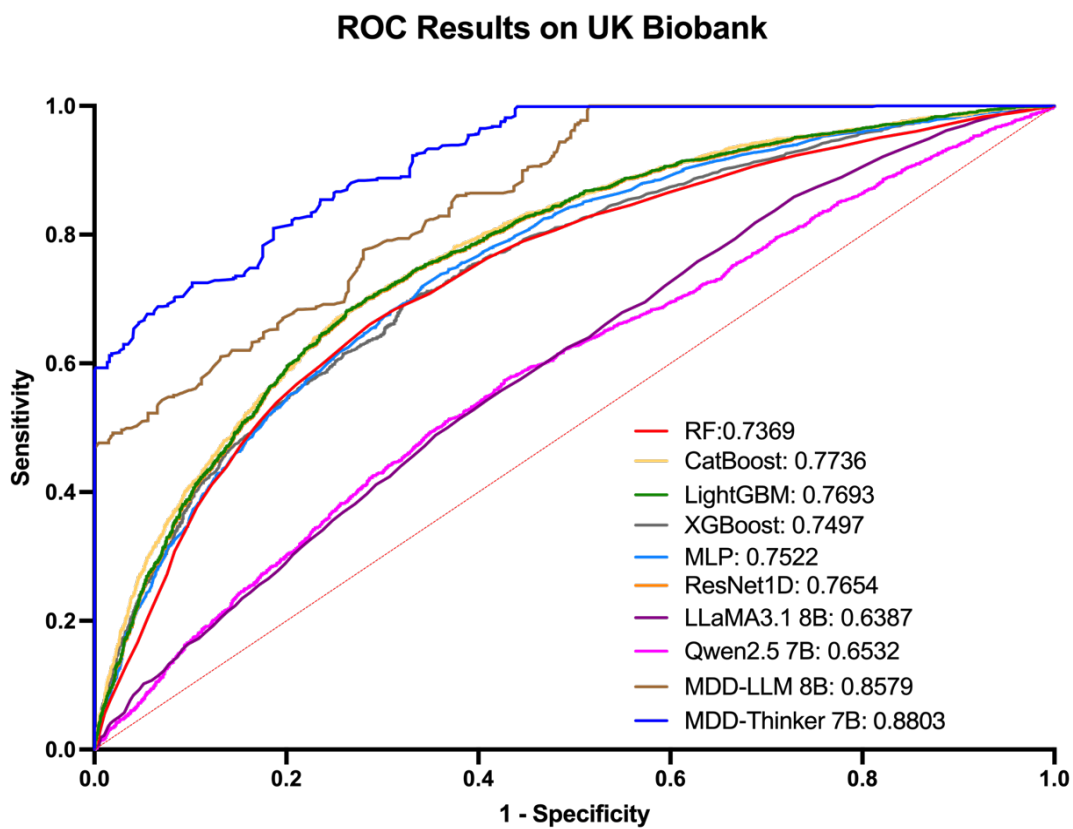

Figure 3. The comparison ROC results of different methods on UK Biobank cohorts.

### 3.4 Effectiveness of SFT and RL

During model fine-tuning, we employed two training strategies, including SFT and RL. Given the complexity of the training process, it was essential to assess whether both approaches contributed positively to the final performance. To this end, we conducted extensive experiments designed to disentangle their respective effects. Specifically, we took the Qwen2.5 7B model as the baseline and evaluated the effects of SFT, RL, and their combination (SFT+RL) on model performance, with experiments carried out on the UK Biobank dataset. The experimental settings were consistent with those described in the Methods section. The results are summarized in Figure 4 and Appendix Table S3, where Figure 4.1 presents radar charts of different methods in terms of multiple evalutation metrics, and Figure 4.2 illustrates performance improvements relative to the baseline. The findings clearly demonstrate that both SFT and RL enhanced model performance. For instance, SFT alone improved accuracy by 22.1% over the baseline. The combined application of SFT and RL achieved the best overall results, yielding relative gains of 29.0%, 38.1%, and 34.8% in accuracy, F1-score, and AUC, respectively. Notably, applying RL directly to the baseline led to moderate improvements, but the effect was less pronounced than that of SFT. This observation highlights the critical role of SFT in strengthening the model's reasoning ability, while RL primarily refines inference processes rather than expanding domain-specific knowledge. These results provide strong evidence that the proposed fine-tuning training strategy substantially improves diagnostic accuracy and generalization for MDD, underscoring the complementary contributions of SFT and RL in optimizing LLM-based clinical reasoning.

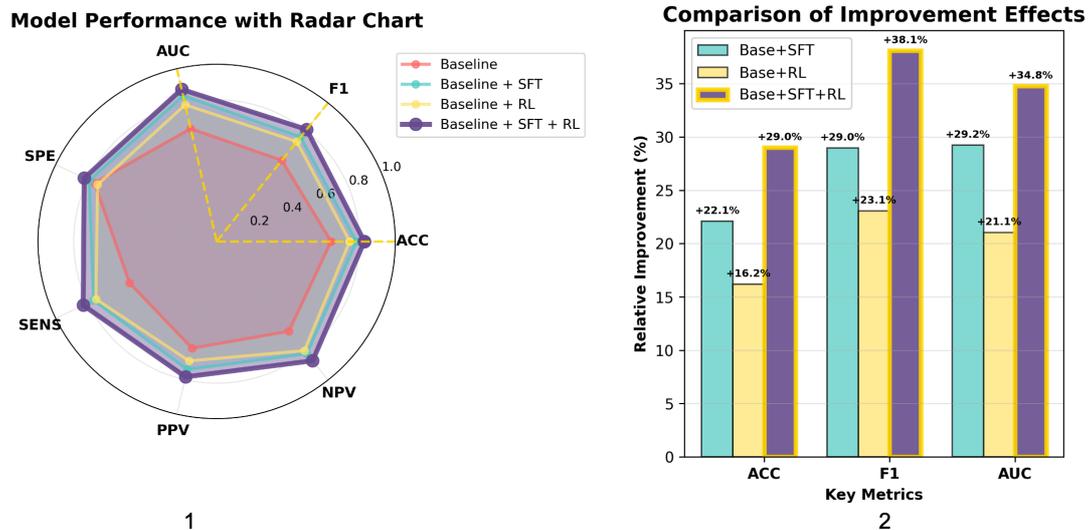

Figure 4. Effectiveness of SFT and RL. Figure 4.1 presents model performance across multiple evaluation metrics using a radar chart, while Figure 4.2 compares the improvement effects achieved by SFT and RL.

### 3.5 Effectiveness of CoT Prompting

CoT prompting, as a straightforward yet effective approach, has attracted substantial attention since the emergence of LLMs. In this study, we integrated CoT prompting as a strategy to improve reasoning ability, resulting in marked enhancements in model performance. However, the complexity of CoT prompts introduces additional computational demands and longer inference times. To systematically examine the impact of CoT prompting on model performance, we designed experiments based on UK Biobank comparing Direct Response, Simple CoT prompting, and Complex CoT prompting. As shown in Table 4, the evaluation metrics included accuracy, F1-score, and the average number of tokens per response. Results indicated that Direct Response produced shorter outputs with limited accuracy. Incorporating Simple CoT prompting significantly improved performance, with the average output length increasing from 36 to 110 tokens. Remarkable, the adoption of Complex CoT prompting yielded the most favorable outcomes. MDD-Thinker not only achieved superior diagnostic accuracy but also generated responses averaging 358 tokens, representing a 9.9 times increase over direct answers and a 3.2 times increase over Simple CoT outputs. These findings demonstrate that the proposed Complex CoT prompting strategy substantially strengthens both reasoning capability and overall model effectiveness.

Table 3. Comparison results of different CoT methods on the UK Biobank dataset.

| Method | Accuarcy | F1-Score | Average Tokens |
| --- | --- | --- | --- |
| Direct Response | 0.6409 | 0.5852 | 36 |
| Simple CoT | 0.7889 | 0.7542 | 110 |
| Complex CoT | 0.8268 | 0.8081 | 358 |

### 3.6 Performance on Public Benchmarks

In this study, MDD-Thinker was primarily developed to advance the diagnosis of MDD. However, the limited availability of publicly accessible MDD-specific datasets makes it difficult to perform a comprehensive comparison with LLM-based soultions. To partially address this limitation, we further evaluated our model on several widely used medical question-answering benchmarks, including MedQA [26], MedMCQA [27], and

PubMedQA [28]. Although MDD-Thinker was not explicitly fine-tuned on large-scale general medical corpora, its knowledge base remains closely connected to the medical domain. The corresponding results are summarized in Table 5. The results demonstrate that MDD-Thinker achieves superior performance compared to models with fewer than 10B parameters, exhibiting competitive capabilities despite its compact architecture. However, when benchmarked against state-of-the-art large-scale systems such as LLaMA-3.1 70B and DeepSeek-R1 70B, performance on open-domain medical datasets shows a notable gap. This disparity can be attributed to two key factors: (1) the inherently limited parameter capacity of our model, and (2) the lack of specialized fine-tuning on extensive medical knowledge corpora. Due to resource limitations, we made a strategic decision to concentrate on MDD diagnosis as a focused clinical application rather than general medical reasoning. Future work will leverage this specialized foundation by integrating comprehensive medical datasets to enhance the model's clinical knowledge breadth and cross-domain applicability.

Table 4. Model performance on public medical benchmarks.

| Method | MedQA | MedMCQA | PubMedQA |
|---|---|---|---|
| BioMistral 7B | 45.0 | 40.2 | 66.9 |
| Mistral 7B | 48.2 | 44.6 | 59.5 |
| LLaMA 3.1 8B | 58.7 | 56.0 | 75.2 |
| Gemma2 9B | 61.8 | 55.9 | 63.3 |
| DeepSeek 67B | 57.1 | 51.7 | 76.1 |
| LLaMA 3.1 70B | 78.4 | 72.5 | 78.5 |
| DeepSeek-R1 70B | 85.6 | 74.3 | 80.0 |
| MDD-Thinker 7B | 59.1 | 56.3 | 74.6 |

## 4. Discussion

In this study, we proposed a RL-based solution to fine-tune LLMs for complex reasoning in MDD diagnosis. To this end, we constructed a reasoning-oriented corpus by integrating advanced LLM outputs with large-scale real-world samples and publicly available data. The MDD-Thinker traning followed a two-stage strategy: first, SFT endowed the model with task-specific knowledge and preliminary reasoning ability for MDD diagnosis; second, RL further refined reasoning performance, leading to improvements in both diagnostic accuracy and reasoning precision.

To evaluate the effectiveness of the proposed MDD-Thinker, we conducted comparisons with classical machine learning algorithms, deep learning models, and existing LLM-based methods. Experimental results demonstrate that our method achieves superior diagnostic performance. Additionally, ablation studies on the SFT and RL training stages indicate that incorporating RL significantly enhances the model's diagnostic and reasoning capability. Although testing on public datasets did not reach state-of-the-art performance, our approach still achieved competitive results among current advanced LLMs solutions. Notably, since our method does not rely on extensive specialized biomedical informatics data, the model's performance primarily derives from the pre-training stage. We anticipated that incorporating additional relevant training data would further enhance its performance on public datasets.

To further illustrate the model's reasoning ability, we designed a series of experiments to evaluate its performance during the reasoning process. The results indicated that MDD-Thinker produced logical, coherent, and interpretable reasoning paths, which systematically led to accurate diagnostic conclusions. These findings not only validated the effectiveness of our approach but also demonstrated the potential of large language models in addressing complex medical reasoning tasks. Moreover, as illustrated in Figure 5, the deployed model consistently maintained high reasoning quality in real-world scenarios, underscoring its practical applicability.

## 5. Ethical Implications

In recent years, LLMs based approaches have shown considerable potential in mental health, particularly for MDD. LLMs can extract information from electronic health records, patient reports, online questionnaires, or social media texts, achieving high accuracy in symptom detection and risk prediction while providing reasoning chains to support clinical decisions. However, their practical application raises critical ethical concerns. Patient privacy and data security are paramount because MDD diagnosis involves sensitive personal information, including medical history, psychological state, and behavioral characteristics [29]. Improper handling during training or inference may lead to breaches or misuse, requiring strict compliance with data protection standards. Model interpretability and transparency are also essential. Outputs without clear reasoning could result in misdiagnosis or inappropriate interventions. Bias and fairness must be addressed, as disparities in training data may cause uneven performance across gender, age, cultural, or socioeconomic groups and potentially exacerbate healthcare inequalities [30]. Finally, responsibility and clinical oversight are crucial, LLMs should serve as decision support tools, and all outputs must be reviewed by clinicians to ensure accountability and patient safety.

## 6. Limitation and Future Work

In this study, we developed an LLM-based framework for diagnosing MDD that combines accurate diagnosis with complex reasoning and interpretable outputs. Despite its contributions, this study is subject to several important limitations. First, the model was primarily trained and evaluated on the UK Biobank dataset, a large-scale resource that lacks sufficient diversity in terms of race, age, and cultural background, potentially limiting the generalizability of the findings. Second, only a subset of publicly available datasets was incorporated due to computational and resource constraints. While these data provided general medical knowledge, their limited coverage may have constrained diagnostic specificity for MDD. Third, current LLMs face challenges in continual learning, as integrating new data typically requires full retraining, leading to high computational costs and reduced adaptability in dynamic clinical environments.

Future research should focus on collecting multi-center, multi-ethnic, and international high-quality MDD datasets to improve generalization. Optimized training strategies and data generation methods should be explored to integrate specialized medical knowledge effectively. Retrieval-augmented generation (RAG) approaches may allow dynamic knowledge updating without frequent retraining. Further evaluation of model interpretability, decision safety, and collaboration with clinicians is essential to advance practical applications in mental health care.

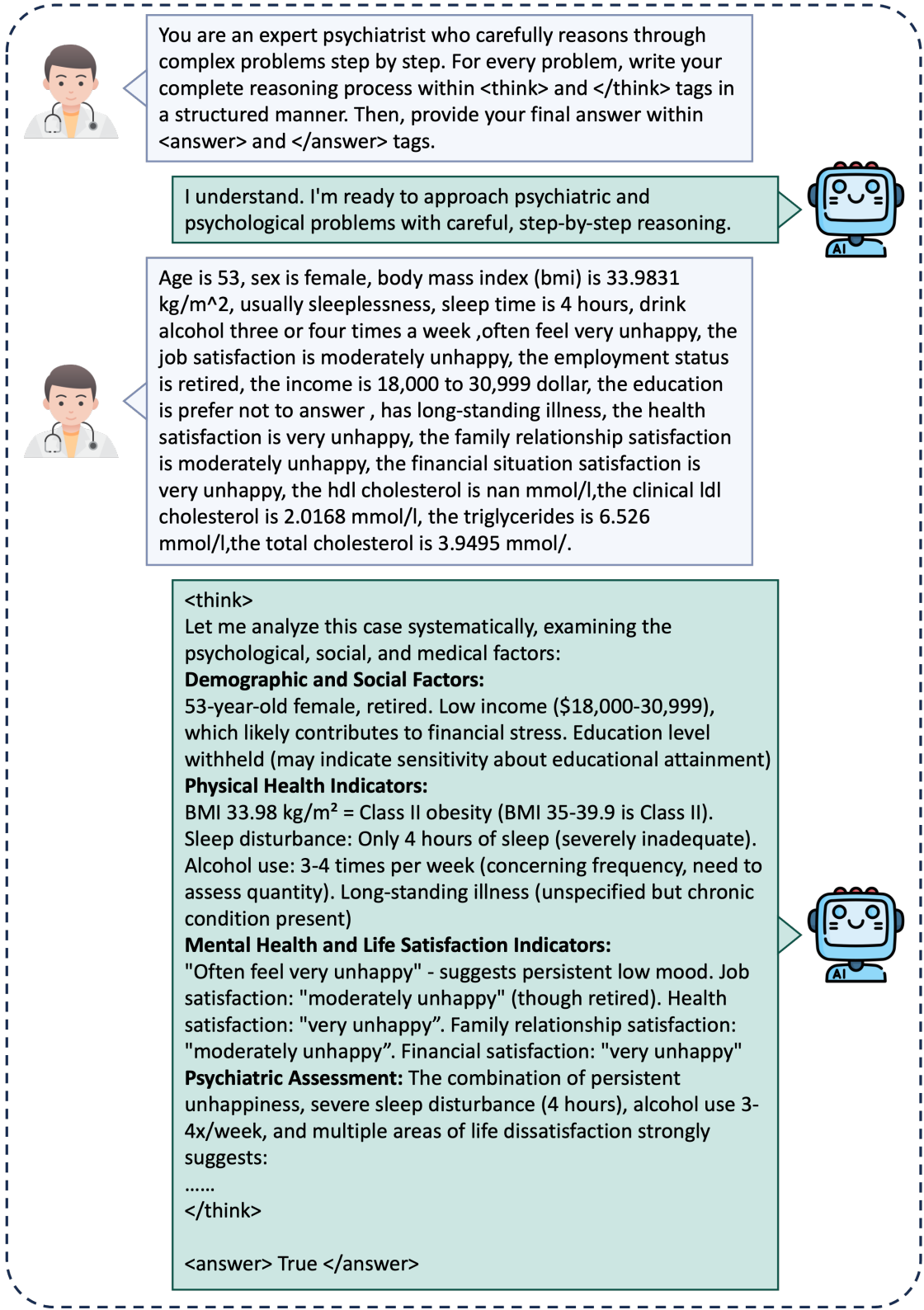

Figure 5. Example showing the capability of MDD-Thinker in model interpretability, reasoning, and dialogue capability.

**Author Contribution**

Concept and design: Y. Sha, K. Li. Acquisition, analysis, or interpretation of the data: Y. Sha, H. Pan, G. Luo, C. Shi, J. Wang. Code and model build: Y. Sha. Drafting of the manuscript: Y. Sha. Critical revision of the manuscript for important intellectual content: K. Li. All the authors discussed the results and commented on the manuscript. All authors agree to be accountable for all aspects of work ensuring integrity and accuracy.

**Availability of Data and Materials**

The source code used and/or analyzed during the current study is available through GitHub (https://github.com/syysha0k/MDD-Thinker) for research purpose only. This research has been conducted using the UK Biobank Resource under Application Number 99946.

**Declaration of Competing Interest**

The authors declare that they have no conflicts of interest.

**Funding**

This work was supported by the grants from Macao Polytechnic University (RP/FCA-14/2023), The Science and Technology Development Funds (FDCT) of Macao (0033/2023/RIB2), and Macau Science and Technology Development Fund and the Department of Science and Technology of Guangdong Province (FDCT-GDST, 0009/2024/AGJ). The funders have no role in study design, data collection and analysis.